\pdfoutput=1

\documentclass[11pt, a4paper]{article}

\usepackage[hyperref]{acl2021}

\aclfinalcopy 

\usepackage{times}
\usepackage{latexsym}

\usepackage[T1]{fontenc}

\usepackage[utf8]{inputenc}

\usepackage{microtype}

\usepackage{inconsolata}

\usepackage{amsmath}
\usepackage{amsfonts}
\usepackage{booktabs}
\usepackage{multicol}
\usepackage{subcaption}
\usepackage{comment}

\usepackage[export]{adjustbox}

\usepackage{xspace}
\usepackage{adjustbox}

\newcommand{\smatch}[0]{\textsc{Smatch}\xspace}

\newcommand{\smaragdtitle}[0]{SMARAGD\includegraphics[angle=10,scale=0.010, trim={0cm 4cm 0cm 0cm}, clip]{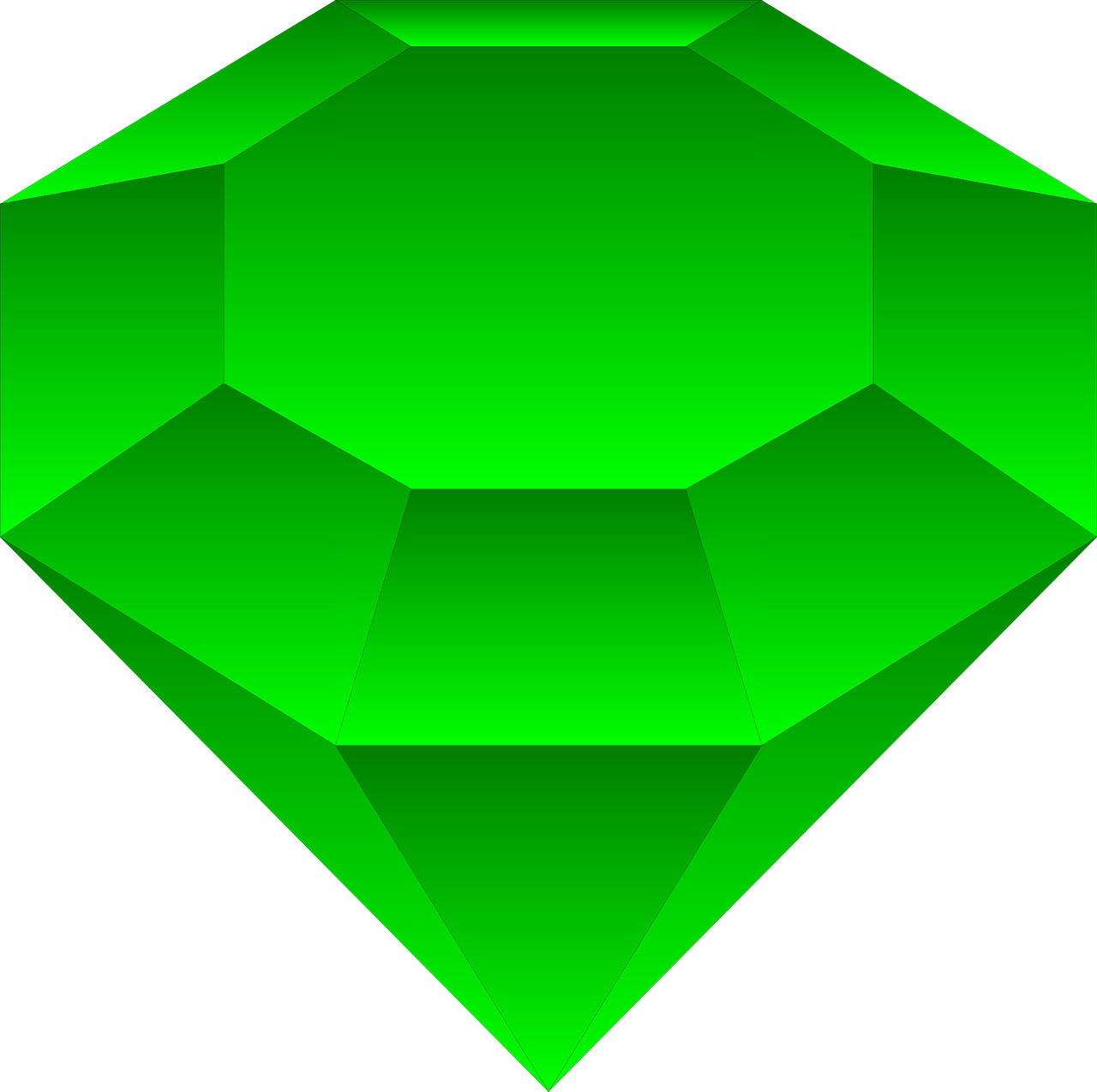}}
\newcommand{\smaragd}[0]{SMARAGD\includegraphics[angle=10,scale=0.007, trim={0cm 4cm 0cm 0cm}, clip]{pics/diamond-g8e5856b6d_1280.png}\xspace}

\title{\smaragdtitle: Learning SMatch for Accurate and Rapid Approximate Graph Distance}

    \author{Juri Opitz ~~~ Philipp Meier ~~~ Anette Frank \\
  Dept.\ of Computational Linguistics \\
  Heidelberg University \\
  69120 Heidelberg \\
 {\tt $\lbrace$opitz,meier,frank$\rbrace$@cl.uni-heidelberg.de} }

\begin{document}
\maketitle
\begin{abstract}
The similarity of graph structures, such as Meaning Representations (MRs), is often assessed via structural matching algorithms, such as \smatch \cite{cai-knight-2013-smatch}. However, \smatch involves a combinatorial problem that suffers from NP-completeness, making large-scale applications, e.g., graph clustering or search, infeasible. To alleviate this issue, we learn \smaragd: Semantic Match for Accurate and Rapid Approximate Graph Distance. We show the potential of neural networks to approximate \textsc{Smatch} scores, i) in linear time using a machine translation framework to predict alignments, or ii) in constant time using a Siamese CNN to directly predict \textsc{Smatch} scores. We show that the approximation error can be substantially reduced through data augmentation and graph anonymization.
\end{abstract}

\section{Introduction}

Semantic graphs such as Meaning Representation (AMR) are directed, rooted and acyclic, and labeled. For instance, in AMR \cite{banarescu-etal-2013-abstract} labels indicate the events and entities of a sentence, and structures capture semantic roles and other key semantics such as coreference.

Often, pairs of MRs need to be  studied, using MR metrics.  Classically, MRs are compared to assess Inter Annotator Agreement in SemBanking or for the purpose of parser evaluation, typically using the \textit{structural} \smatch metric \cite{cai-knight-2013-smatch, opitz-2023-smatch}. Going beyond these applications, researchers have leveraged \smatch-based MR metrics for NLG evaluation \cite{opitz2020towards, manning-schneider-2021-referenceless}, for re-inforcing AMR parsers \cite{naseem-etal-2019-rewarding}, as a  basis for a COVID-19 semantics-based search engine \cite{bonial-etal-2020-infoforager}, comparison of cross-lingual AMR \cite{uhrig-etal-2021-translate, wein-etal-2022-effect}, and fine-grained argument similarity assessment \cite{opitz-etal-2021-explainable}. Many of these extended scenarios greatly profit from a \textit{quick similarity computation}. Also, additional future applications can be anticipated that require fast metric inference: e.g., corpus linguists who want to find instantiations of abstract semantic patterns in a large corpus.

But graph metrics typically suffer from a high time complexity: Computation of \smatch is NP-hard \cite{nagarajan2009maximum}, and it can take more than a minute to compare some 1,000 AMR pairs \cite{song-gildea-2019-sembleu}. To understand that this can become problematic in many setups, consider a hypothetical user who desires exploring a (small) AMR-parsed corpus with only $n=1,000$ instances via clustering. The (symmetric) \smatch needs to be executed over $(n^2-n)/2 =499,500$ pairs, resulting in a total time of more than 6 hours. 

This high time complexity is a well-known bottleneck and negatively impacts AMR evaluation time \cite{song-gildea-2019-sembleu}, as well as parsing efficency of approaches involving re-inforcement learning \cite{naseem-etal-2019-rewarding} or graph ensembling \cite{NEURIPS2021_479b4864}, where the \smatch metric is executed with high frequency. Furthermore, given recent interest into larger meaning representations that cover multiple sentences, such as multi-sentence AMR \cite{ogorman-etal-2018-amr}, dialogue AMR \cite{bonial-EtAl:2021:IWCS} or discourse representation structures \cite{kamp1981theory, van-noord-etal-2018-evaluating}, we anticipate that this problem will become more pressing in the future. 

Testing ways to mitigate these issues, we propose a method that learns to match semantic graphs from a teacher \smatch, and show that this can reduce AMR clustering time from hours to seconds, with only little expected loss in accuracy.

Our contributions are:

\begin{enumerate}
    \item We explore three different neural approaches to synthesize the combinatorial graph metric \smatch from scratch.
    \item We show that we can approximate \smatch up to a small error, by leveraging novel data augmentation tricks. 
\end{enumerate}

Our code is available at: \url{https://github.com/PhMeier/Smaragd/}. 

\section{Related work}

\paragraph{Other metrics for MR similarity} Recently, researchers have proposed AMR metrics beyond \smatch. We can distinguish two lines of work: i) metrics aiming at extreme efficiency by skipping the alignment and extracting graph parts via breadth-first traversal \cite{song-gildea-2019-sembleu, anchieta2019sema}. ii) Weisfeiler-Leman graph metrics that aim to reflect human similarity ratings \cite{opitz2021weisfeiler}. \citet{opitz-tacl} make an argument for the importance of graph alignment.

\paragraph{Algorithm synthesis} Neural networks have been studied for solving other problems efficiently. Examples range from sorting numbers \cite{graves2014neural, DBLP:journals/corr/NeelakantanLS15} to solving elaborated tasks such as symbolic integration \cite{lample2019deep}, the famous traveling salesman  problem \cite{gambardella1995ant, budinich1996self, bello2016neural, zhang2021solving}, and computer programs \cite{DBLP:journals/corr/BalogGBNT16, NEURIPS2020_7a685d9e, NEURIPS2021_ba3c95c2}. The `long-range arena' benchmark \cite{tay2021long} includes algorithm synthesizing tasks, such as `listOps' (learning to calculate), or Xpath (tracing a squiggly line), which prove challenging even for SOTA architectures. Since structural graph matching with \smatch constitutes a very hard combinatorial problem, investigating efficient neural approximations seems an interesting challenge in general -- beyond the use-case of rapid graph distance calculation.

\section{Learning NP-hard graph alignment}

The \smatch metric measures the structural overlap of two graphs. We i) compute an alignment between variable nodes of graphs and ii) assess triple matches based on the provided alignment. Formally, we start with two graphs $a$ and $b$ with variable nodes $X = (x_1,...x_n)$ and $Y=(y_1...y_m)$. The goal is then to find an optimal \textit{alignment} 
\begin{equation}
    \label{eq:map}
    map^\star: X \rightarrow Y,
\end{equation}
searching for a $map$ that maximizes the number of \textit{triple matches} 
for the two graphs. For instance, assume two AMR triples \texttt{(x, ARG0, y)} $\in \mathcal{G}$ and \texttt{(u, ARG0, v)} $\in \mathcal{G}'$. If $x = u$ and $y = v$, we count \textit{one} triple match. Finally:
 
\begin{equation}
    \label{eq:score}
    SMATCH = \max_{map} score(a, b, map)
\end{equation}

Researchers typically use a harmonic mean based overlap $score= F1 = 2PR/(P+R)$, where  $P = |triples(a) \cap triples(b)| / |triples(a)$ and $R = |triples(a) \cap triples(b)| / |triples(b|$. 
 
\subsection{Setup}

\paragraph{Experimental data creation} We create the data for our experiments as follows: 1. We parse 59,255 sentences of the  LDC2020T02 AMR dataset with a parser \cite{lyu-titov-2018-amr} to obtain graphs that can be aligned to reference graphs; 2. For every parallel graph pair $(a, b)$, we use \smatch (ORACLE) to compute an F1 score $s$ and the alignment $map^\star$, yielding an extended data tuple $(a,b,s,map^\star)$ We shuffle the data and split it into training, development and test set (56255-1500-1500).

\paragraph{Objective and approach} The task is to reproduce the teacher ORACLE as precisely as possible. We design and test three different approaches. The first is indirect, in that it predicts the alignment, from which we compute the score. The second directly predicts the scores. The third approach enhances the second, to make it even more efficient.

\begin{figure*}[h]
    \centering
    \includegraphics[width=0.8\linewidth]{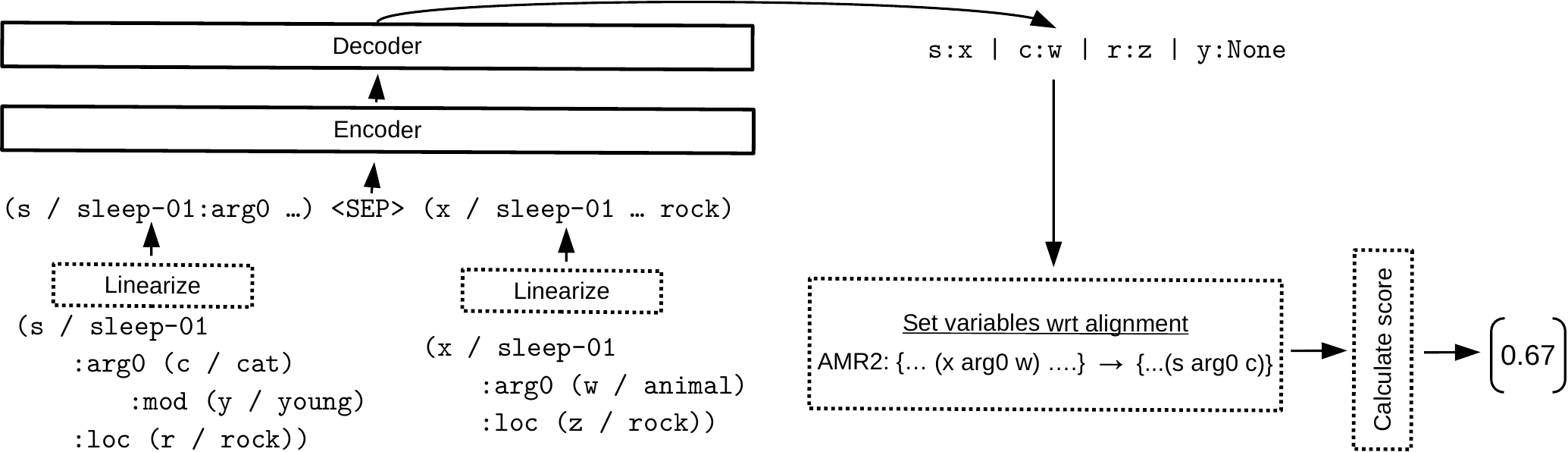}
    \caption{Seq2seq \smatch alignment-learner.}
    \label{fig:s2s}
\end{figure*}
\begin{figure*}[h]
    \centering
    \includegraphics[width=0.8\linewidth]{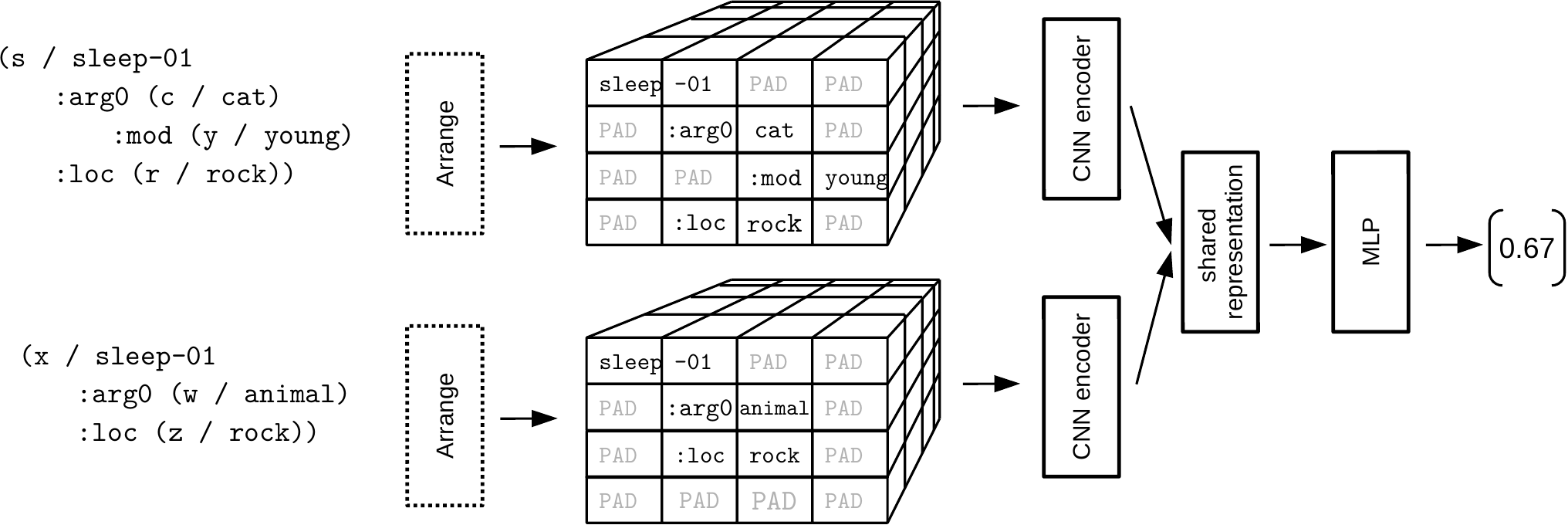}
    \caption{Implicit CNN-based \smatch graph metric 
    predictor.}
    \label{fig:cnn}
\end{figure*}

\subsection{Synthesis option I: Alignment learning}

Here, we aim to learn the alignment itself (Eq.\ \ref{eq:map}) with an NMT model, as illustrated
in Figure \ref{fig:s2s}. For the input, we linearize the two AMRs and concatenate the linearized token sequences with a special \texttt{<SEP>} token. The output consists of a sequence \textit{$x_j$:$y_k$ ... $x_i$:$y_m$ ...} where in every pair $u$:$v$, $u$ is a variable node from the first AMR mapped to a node $v$ from the second AMR. The \smatch score is then calculated based on the predicted alignment.

To predict the node alignments/mapping of variables, we use a transformer based encoder-decoder NMT model. Details about the network structure and hyperparameters are stated in Appendix \ref{app:seq2seq}.

\subsection{Synthesis option II: \smatch prediction} 

In this setup, we aim to predict \smatch F1 scores for pairs of AMRs directly, in a single step. This means that we  directly learn Eq.\ \ref{eq:score} with a neural network and our target is the ORACLE F1 score. 

To learn this mapping, we adapt the convolutional neural network (CNN) of \citet{opitz-2020-amr}, as shown in Figure \ref{fig:cnn}. 
The model was originally intended to assess AMR accuracy \cite{opitz-frank-2019-automatic}, i.e., measuring AMR parse quality without a reference. 
Taking inspiration from human annotators, who exploit a spatial `Penman' arrangement of AMR graphs for better understanding, it models directed-acyclic and rooted graphs as 2d structures, employing a CNN for processing, which is highly efficient. To feed a pair of AMRs, we remove the dependency graph encoder of the model and replace it with the AMR graph encoder. Moreover, we increase the depth of the network by adding one more MLP layer after convolutional encoding. A basic mean squared error is employed as loss function. More details about hyperparameters are stated in Appendix \ref{app:cnn}. 

\subsection{Synthesis option III: AMR Vector learning}

Inspired by \citet{reimers-gurevych-2019-sentence}, we aim to make the CNN even more efficient, by alleviating the need for pair-wise model inferences. Instead of computing a shared representation of two CNN-encoded graphs, we process each representation with 
an MLP (w/ shared parameters), to obtain two vectors $NN(a)$ and $NN(b)$. These vectors are then tuned with signal from ORACLE($s$): 
\begin{equation}
    \mathcal{L} = \sum_{(a, b, s)} \bigg(\big[1 - |NN(a) - NN(b)|\big] - s\bigg)^2,
\end{equation}

where $||$ is returns a vector distance $\in [0,1]$. This approach enables extremely fast search and clustering: the required (clustering-)model inferences are $O(n)$ instead of $O(n^2)$, since the similarity is achieved with simple linear vector algebra.

\subsection{Data compression and extension tricks}

\begin{figure}
    \centering
    \includegraphics[width=0.8\linewidth]{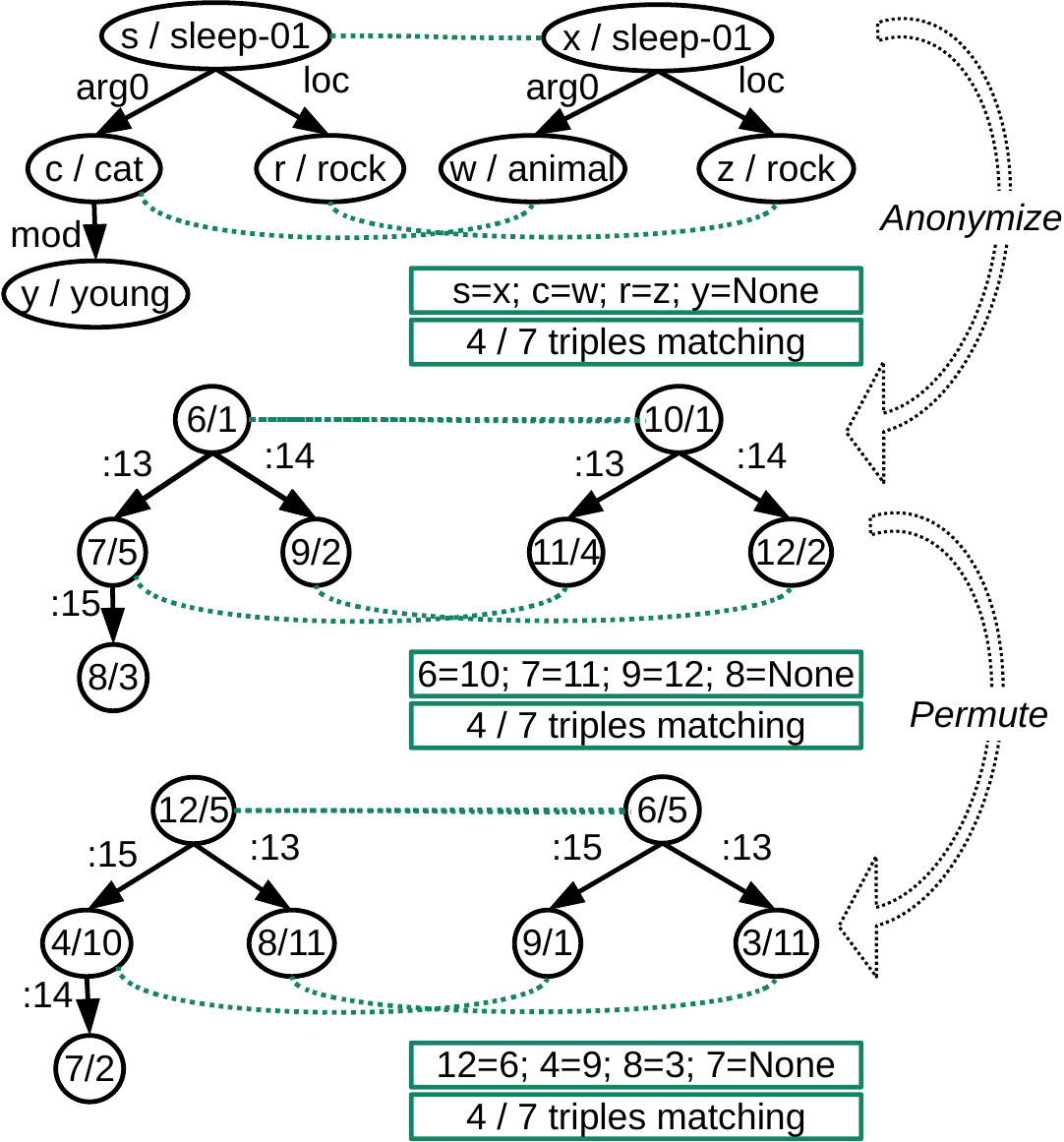}
    \caption{AMR graph anonymization and permutation.}
    \label{fig:anon}
\end{figure}

\paragraph{Vocabulary reduction trick} 
The \smatch metric measures the structural overlap of two graphs. This means that we can greatly reduce our vocabulary, by assigning each graph pair a \textit{local vocabulary} (see Figure \ref{fig:anon}, `anonymize'). 

First, we gather all nodes from two graphs $a$ and $b$, computing a joint vocabulary over the concept nodes. We then relabel the concepts with integers starting from 1. E.g., consider AMR $a$: \textit{(r / run-01 :ARG0 (d / duck))}, and AMR $b$: \textit{(x / run-01 :ARG0 (y / duck) :mod (z / fast))}. The gold alignment is $map^\star =\{(r,x), (d,y), (\emptyset, z)\}$. Now, we set the shared concepts and relations to the same index \textit{run=run=1} and \textit{duck=duck=2} and \textit{:ARG0=:ARG0=3} and distribute the rest of the indices \textit{r=4, d=5, x=6, y=7, z=8, fast=9, :mod=10}. This yields equivalent AMRs $a'$ = \textit{(4 / 1 :3 (5 / 2))} and $b'$ = \textit{(6 / 1 :3 (7 / 2) :10 (8 / 9))}. The target alignment then 
equals $map^\star =\{(4,6), (5,7), (\emptyset, 8)\}$. This strategy greatly reduces the vocabulary size, in our case from 40k tokens to less than 700.

\paragraph{Auxiliary data creation trick} We also find that we can cheaply create auxiliary gold data. We re-assign different indices to AMR tokens, and correspondingly modify the ORACLE alignment (Figure \ref{fig:anon}, `permute'). In our experiments, we permute the existing token-index vocabularies 10 times, resulting in a ten-fold increase of the training data. We 
expect that, with this strategy, the model will better learn properties of permutation invariance, which in turn will help it synthesize the algorithm.

\subsection{Evaluation} 

\paragraph{Output post-processing} For the score synthesis (Option II) and vector synthesis (Option III), no further post-processing is required, since we directly obtain the estimated  \smatch scores as output. In the explicitly synthesized alignment algorithm, however, we get $map$, which is the predicted alignment from the sequence-to-sequence model. In this case, we simply feed $map$ as an argument into Eq.\ \ref{eq:score}, to obtain the scores.

\paragraph{Evaluation} We compare the predicted scores $\hat{y}$ against the gold scores $y$ with Pearson's $\rho$. However, for the model that predicts the explicit alignment (Option I), we can compute another interesting and meaningful metric. For this, we first calculate the average \smatch score over AMR pairs given the gold alignment $map^\star$, and then we calculate the average \smatch score over AMR pairs given the predicted alignment $\widehat{map}$ using Eq.\ \ref{eq:score}. Note, that the \smatch score based on the gold alignment constitutes an upper bound (max). Therefore, the \smatch score based on the predicted alignment shows us how close we are to this upper bound. Our baseline consists of scores that are computed from a random alignment (\textit{random}).

\begin{table}
    \centering
    \scalebox{0.78}{\begin{tabular}{llrrrrr}
   & data trick &  Eq.\ \ref{eq:score}  & Pea's $\rho$ & time$^{(secs)}$ \\
    \midrule
        ORACLE & na & 77.5  & 100 & 28680 \\
        rand.\ baseline & na & 13.5 &   22.2 & 0.4\\
        \midrule
        \midrule
       align.\ synthesis& &  39.0   & 52.8 & 1089\\
       align.\ synthesis & voc & 64.5   & 80.0 & 1089\\
       align.\ synthesis & voc+aug & 76.4 &  \textbf{98.4} & 1089 \\
        \midrule
        score synthesis & &na  &\underline{87.5} & 140 \\
        score synthesis & voc & na  & \textit{82.0} &  140\\
        score synthesis & voc+aug& na & 96.8 & 140 \\
        \midrule
        vector synthesis & & na &  84.7 & 0.7\\
        vector synthesis  &voc & na  & 75.6 & 0.7\\
        vector synthesis &voc+aug& na & 94.2 & 0.7 \\
        \bottomrule
    \end{tabular}}
    \caption{Results of experiments. time: Approximate time for computing a pair-wise distance matrix on 1k AMRs on a TI 1080 GPU.}
    \label{tab:res-table}
\end{table}

\paragraph{Results (Table \ref{tab:res-table})} Our best model is the NMT approach using both data augmentation tricks. Obtaining 98.4 $\rho$, it very closely approximates the ORACLE, while being about 30 times faster than ORACLE and 76.2 points better then the random baseline. Perhaps the best tradeoff between speed and approximation performance is gained by the simple CNN score synthesis (96.8 $\rho$, 200x faster than ORACLE), also using both data tricks. The vector synthesis falls a bit shorter in performance (94.2 $\rho$), but it is extremely fast and achieves a 40,000x speed-up compared to ORACLE and about 1500x compared to the NMT approach.\footnote{Note also that all models in Table \ref{tab:res-table} are significantly better (p$<$0.001) than the random baseline (one-sided test w/ z-transform).}

Consistently, the data extension (\textit{aug}) is very useful. However, the vocabulary reduction (\textit{voc}) is only useful for the NMT model (+27.2 points), whereas the scores are lowered for the CNN-based models ($-$5.5 for score synthesis, $-$9.1, \textit{vector synthesis}). We conjecture that the CNNs learn \smatch more indirectly by exploiting token similarities in the global vocabulary, and therefore struggle more to build a generalizable algorithm, in contrast to the bigger NMT transformer that learns to assess tokens fully from their given graph context.

\section{Conclusion}

We tested methods for learning to solve the hard structural graph matching problem that is key to many applications where we compare meaning representations. To this aim, we explored different neural architectures, and data augmentation strategies that help models to generalize. Our best models increase metric calculation speed by a large factor while incurring only small losses in accuracy that can be tolerated in many use cases. Our work paves the way to emergent use-cases of meaning representation that involve pair-wise analysis: e.g., semantic clustering or semantic pattern-based search for corpus linguistic studies.

\section*{Limitations}

An issue of the tested methods concerns the alignment of larger graphs with many variables. On one hand, when the alignment candidate space increases, the runtime of \smatch increases exponentially, while our considered approaches remain fast. However, in such a scenario, the neural models are bound to trade in some accuracy. Table \ref{tab:accuracyvsvar} (Appendix \ref{app:problemsizes}) assesses the effect size for differently sized alignment candidate spaces: while the model overall copes with different search space sizes, the accuracy loss is more considerable for large problems. We conclude that the fast and accurate alignment of \textit{larger} AMR graphs remains a challenging and unsolved problem. However, note that such a bottleneck even exists for the algorithmic metrics, which either use a hill-climber that suffers from worsening sub-optimality or require a costly ILP procedure that may be infeasible for larger graphs (see \citet{opitz-2023-smatch} for discussion and analysis). In this regard, we believe that our proposed data extension trick in combination with long-sequence transformers \cite{beltagy2020longformer, Rae2020Compressive, choromanski2021rethinking} may provide valuable means to address this limitation, or provide useful tradeoffs.

Other limitations are: i) the models that were trained without our proposed anonymization protocol were tested on graphs that contain English concepts, and therefore depend on an English vocabulary. ii) For loading the models, our tested methods require more RAM memory than \smatch, which can be calculated on a low-budget computer.

\bibliography{acl2021}
\bibliographystyle{acl_natbib}

\appendix
\newpage
\section{Appendix}
\label{sec:appendix}

\subsection{Sequence-to-sequence network parameters}
\label{app:seq2seq}
 \begin{table}
     \centering
     \begin{tabular}{ll}
     \toprule
         parameter & value \\
         \midrule
        embedding size &  512  \\
        encoder & 4 transformer layers w/ 4 heads \\
        decoder & 4 transformer layers w/ 4 heads \\
        feed forw.\ dim & 2048 \\
        loss & cross-entropy \\
        weight init & xavier \\
        optimizer & adam \\
        learning rate & 0.0002 \\
        batch size & 8192 (tokens) \\
        \bottomrule
     \end{tabular}
    \caption{Overview of NMT hyper-parameters.}
    \label{tab:nmthps}
 \end{table}

Hyper-parameters for the NMT approach are displayed in Table \ref{tab:nmthps}. The best model is determined on the development data by calculating BLEU against the reference alignments. 

\subsection{CNN network parameters}
\label{app:cnn}

\begin{table}
    \centering
    \scalebox{0.9}{
    \begin{tabular}{ll}
    \toprule
         parameter& value \\
         \midrule
         emb.\ dimension & 100 \\
         `pixels' & 60x15 \\
        CNN encoder & concatenate( \\
        & 256 3x3 convs, 3x3 max pool \\
        & 128 5x5 convs, 5x5 max pool) \\
        MLP & relu layer followed by lin.\ regressor \\
        weight init & xavier \\
        optimizer & adam \\
        learning rate & 0.001 \\
        batch size & 64 \\
         \bottomrule
    \end{tabular}}
    \caption{Overview of CNN hyper-parameters.}
    \label{tab:cnnhps}
\end{table}

Hyper-parameters for the CNN approach are displayed in Table \ref{tab:nmthps}. The best model is determined on the development data by calculating Pearson's $\rho$ correlation of predicted scores and gold scores.

\subsection{Analysis of performance on different problem sizes}
\label{app:problemsizes}

See Table \ref{tab:accuracyvsvar}.
\begin{table}
    \centering
    \scalebox{0.88}{
    \begin{tabular}{lllll}
    \toprule
    & &\multicolumn{2}{c}{$\Delta$ vs.\ ORACLE} \\
         data type & data size &Eq.\ \ref{eq:score} & Pea's $\rho$ & better\\
         \midrule
         full & 1500 & -1.1 & -1.6 & - \\
         \midrule
         $<5$ vars & 505 &  -0.6 & -1.2 & yes \\
        $<10$ vars & 1041 &  -0.7 & -1.2 & yes \\
        $<15$ vars & 1206 &  -0.9 & -1.2 & yes \\
        $<20$ vars & 1353 &  -0.9 &-1.2 & yes \\
        $<25$ vars & 1449 &  -0.9 &-1.3 & yes \\
        \midrule
         $>5$ vars & 940 & -1.5 & -2.2 & no \\
        $>10$ vars & 476 & -2.1 & -3.6 & no\\
        $>15$ vars & 318 & -2.3 & -5.4 & no\\
        $>20$ vars & 183 & -3.0 & -10.1& no\\
        $>25$ vars & 83 & -4.7 & -19.3 & no\\
        $>30$ vars & 37 & -8.0 & -25.5 & no\\
        $>35$ vars & 20 &  -12.5 & -41.1 & no\\
        \midrule
        single snt AMRs & 1421& -1.0 & -1.5 & yes \\
        multi snt AMRs & 79 & -2.7 & -9.6 & no\\

         \bottomrule
    \end{tabular}}
    \caption{Experiments on different test subsets that represent different problem complexities predicted with our best model (\textit{align. synthesis+voc+aug}). $<> x$ vars means that one of two graphs contains $<> x$ variables. \textit{better}: is the drop in accuracy of the model vs. ORACLE smaller compared with the model tested on all data?}
    \label{tab:accuracyvsvar}
\end{table}

\end{document}